\documentclass[runningheads]{llncs}

\usepackage[english]{babel}
\usepackage[utf8]{inputenc}
\usepackage[T1]{fontenc}
\usepackage{booktabs,multirow}
\usepackage[table]{xcolor}
\usepackage[colorinlistoftodos]{todonotes}
\usepackage{cite}
\usepackage{hyperref}
\usepackage{float} 
\begin{document}
%
\title{The Benefits of Close-Domain Fine-Tuning for Table Detection in Document Images\thanks{This work was partially supported by Ministerio de Econom\'ia y Competitividad [MTM2017-88804-P], Ministerio de Ciencia, Innovación y Universidades [RTC-2017-6640-7], Agencia de Desarrollo Econ\'omico de La Rioja [2017-I-IDD-00018], and the computing facilities of Extremadura Research Centre for Advanced Technologies (CETA-CIEMAT), funded by the European Regional Development Fund (ERDF). CETA-CIEMAT belongs to CIEMAT and the Government of Spain.}}
\titlerunning{Close-Domain Fine-Tuning for Table Detection in Document Images}


\author{\'Angela Casado-Garc\'ia\inst{1} \and C\'esar Dom\'inguez\inst{1}\orcidID{0000-0002-2081-7523} \and J\'onathan Heras\inst{1}\orcidID{0000-0003-4775-1306} \and Eloy Mata\inst{1}\orcidID{0000-0003-0538-4579} \and Vico Pascual\inst{1}\orcidID{0000-0003-3576-0889}}
\authorrunning{A. Casado-Garc\'ia et al.}
\institute{Department of Mathematics and Computer Science, University of La Rioja, Spain
\email{\{angela.casado, cesar.dominguez, jonathan.heras, eloy.mata, vico.pascual\}@unirioja.es}}

\maketitle

\begin{abstract}
A correct localisation of tables in a document is instrumental for determining their structure and extracting their contents; therefore, table detection is a key step in table understanding. Nowadays, the most successful methods for table detection in document images employ deep learning algorithms; and, particularly, a technique known as \emph{fine-tuning}. In this context, such a technique exports the knowledge acquired to detect objects in natural images to detect tables in document images. However, there is only a vague relation between natural and document images, and fine-tuning works better when there is a close relation between the source and target task. In this paper, we show that it is more beneficial to employ fine-tuning from a closer domain. To this aim, we train different object detection algorithms (namely, Mask R-CNN, RetinaNet, SSD and YOLO) using the TableBank dataset (a dataset of images of academic documents designed for table detection and recognition), and fine-tune them for several heterogeneous table detection datasets. Using this approach, we considerably improve the accuracy of the detection models fine-tuned from natural images (in mean a 17\%, and, in the best case, up to a 60\%).  

\keywords{Table detection \and Deep learning \and Transfer learning \and Fine-tuning}
\end{abstract}

\section{Introduction}

Tables are widely present in a great variety of documents such as administrative documents, invoices, scientific papers, reports, or archival documents among others; and, therefore, techniques for table analysis are instrumental to automatically extract relevant information stored in a tabular form from several sources~\cite{Couasnon14}. The first step in table analysis is \emph{table detection} --- that is, determining the position of the tables in a document --- and such a step is the basis to later determine the internal table structure and, eventually, extract semantics from the table contents~\cite{Couasnon14}. 

Table detection methods in digital born documents, such as readable PDFs or HTML documents, employ the available meta-data included in those documents to guide the analysis by means of heuristics~\cite{Oro09}. However, table detection in image-based documents, like scanned PDFs or document images, is a more challenging task due to high intra-class variability --- that is, there are several table layouts that, in addition, are highly dependent on the context of the documents --- low inter-class variability  --- that is, other objects that commonly appear in documents (such as figures, graphics or code listing among others) are similar to tables --- and the heterogeneity of document images~\cite{Embley06}. These three issues make difficult to design rules that are generalisable to a variety of documents; and, this has led to the adoption of machine learning techniques, and, more recently, deep learning methods. 

Currently, deep learning techniques are the state of the art approach to deal with computer vision tasks~\cite{Pyimagesearch}; and, this is also the case for table detection in document images~\cite{tablebank,Schreiber17,Kerwat18}. The most accurate models for table detection have been constructed using \emph{fine-tuning}~\cite{Razavian14}, a \emph{transfer learning} technique that consists in re-using a model trained in a source task, where a lot of data is available, in a new target task, with usually scarce data. In the context of table detection, the fine-tuning approach has been applied due to the small size of table detection datasets that do not contain the necessary amount of images required to train deep learning models from scratch. In spite of its success, this approach has the limitation of applying transfer learning from natural images, a distant domain from document images. This makes necessary the application of techniques, like image transformations~\cite{Gilani17}, to make document images look like natural images.

In this work, we present the benefits of applying transfer learning for table detection from a close domain thanks to the {\LaTeX} part of the TableBank dataset~\cite{tablebank}, a dataset that consists of approximately 200K labelled images of academic documents containing tables --- a number big enough to train deep learning models from scratch. Namely, the contributions of this work are the following:

\begin{itemize}
    \item We analyse the accuracy of four of the most successful deep learning algorithms for object detection (namely, Mask-RCNN, RetinaNet, SSD and YOLO) in the context of table detection in academic documents using the TableBank dataset.
    \item Moreover, we present a comprehensive study where we compare the effects of fine-tuning table detection models from a distant domain (natural images) and a closer domain (images of academic documents from the TableBank dataset), and demonstrate the advantages of the latter approach. To this aim, we employ the 4 aforementioned object detection architectures and 7 heterogeneous table detection datasets containing a wide variety of document images. 
    \item Finally, we show the benefits of using models trained for table detection on document images to detect other objects that commonly appear in document images such as figures and formulas. 
\end{itemize}

As a by-product of this work, we have produced a suite of models that can be employed via a set of Jupyter notebooks (documents for publishing code, results and explanations in a form that is both readable and executable)~\cite{jupyter} that can be run online using Google Colaboratory~\cite{colab} --- a free Jupyter notebook environment that requires no setup and runs entirely in the cloud avoiding the installation of libraries in the local computer. In addition, the code for fine-tuning the models is also freely available. This allows the interested readers to adapt the models generated in this work to detect tables in their own datasets. All the code and models are available at the project webpage \href{https://github.com/holms-ur/fine-tuning}{https://github.com/holms-ur/fine-tuning}.

The rest of this paper is organised as follows. In the next section, we provide a brief overview of the methods employed in the literature to tackle the table detection task. Subsequently, in Section~\ref{sec:methods}, we introduce our approach to train models for table detection using fine-tuning, as well as the setting that we employ to evaluate such an approach. Afterwards, we present the obtained results along with a thorough analysis in Section~\ref{sec:results}, and the tools that we have developed in Section~\ref{sec:tools}. Finally, the paper ends with some conclusions and further work. 

\section{Related work}\label{sec:related-work}

Since the early 1990s, several researchers have tackled the task of table detection in document images using mainly two approaches: rule-based techniques and data-driven methods. The former are focused on defining rules to determine the position of lines and text blocks to later detect tabular structures~\cite{Hirayama95,Jianying99,Zanibbi04}; whereas, the latter employ statistical machine learning techniques, like Hidden Markov models~\cite{Costa09}, a hierarchical representation based on the MXY tree~\cite{Cesari02} or feature engineering together with SVMs~\cite{Kasar13}. However, both approaches have drawbacks: rule-based methods require the design of handcrafted rules, that do not usually generalise to several kinds of documents; and, machine learning methods require manual feature engineering to decide the features of the documents that are feed to machine learning algorithms. These problems have been recently alleviated by using deep learning methods.

Nowadays, deep learning techniques are the state of the art approach to deal with table detection. The reason is twofold: deep learning techniques are robust for different document types; and, they do not need handcrafted features since they automatically learn a hierarchy of relevant features using convolutional neural networks (CNNs)~\cite{Goodfellow16}. Initially, hybrid methods combining rules and deep-learning models were suggested; for instance, in~\cite{Hao16} and~\cite{Borges17}, CNNs were employed to decide whether regions of an image suggested by a set of rules contained a table. On the contrary, the main approach followed currently consists in adapting general deep learning algorithms for object detection to the problem of table detection. Namely, the main algorithm applied in this context is Faster R-CNN~\cite{fasterrcnn}, that has been directly employed using different backbone architectures~\cite{tablebank,Schreiber17,Kerwat18}, and combined with deformable CNNs~\cite{Siddiqui18} or with image transformations~\cite{Gilani17}. Other detection algorithms such as YOLO~\cite{yolov3} or SSD~\cite{ssd} have also been employed for table detection~\cite{Kerwat18,Huang19}, but achieving worse results than the methods based on the Faster R-CNN algorithm. Nevertheless, training deep learning models for table detection is challenging due to the considerable amount of images that are necessary for this task --- up to recently, the biggest dataset of document images containing tables was the Marmot dataset with 2,000 labelled images~\cite{marmot}, far from the datasets employed by deep learning methods that consists of several thousands, or even millions, of images~\cite{ILSVRC15,pascalvoc}.  

In order to deal with the problem of limited amount of data, one of the most successful methods applied in the literature is transfer learning~\cite{Razavian14}, a technique that re-uses a model trained in a source task in a new target task. This is the approach followed in~\cite{Schreiber17,Siddiqui18,Gilani17}, where they use models trained on natural images to fine-tune their models for table detection. However, transfer learning methods are more effective when there is a close relation between the source and target domains, and, unfortunately, there is only a vague relation between natural images and document images. This issue has been faced, for instance, by applying image transformations to make document images as close as possible to natural images~\cite{Gilani17}.

Another option to tackle the problem of limited data consists in acquiring and labelling more images, a task that has been undertaken for table detection in the TableBank project~\cite{tablebank} --- a dataset that consists of 417K labelled images of documents containing tables. The TableBank dataset opens the door to apply transfer learning to not only construct models for table detection in different kinds of documents, but also to detect other objects, such as figures or formulas, that commonly appear in document images. This is the goal of the present work.

\section{Materials and methods}\label{sec:methods}
In this section, we explain the fine-tuning method, as well as the object detection algorithms, datasets and evaluation metrics used in this work.

\subsection{Fine-tuning}
Transfer learning allows us to train models using the knowledge learned by other models instead of starting from scratch. The idea on which transfer learning techniques are based is that CNNs are designed to learn a hierarchy of features. Specifically, the lower layers of CNNs focus on generic features, while the final ones focus on specific features for the task they are working with. As explained in~\cite{Razavian14}, transfer learning can be employed in different ways, and the one employed in this work is known as fine-tuning. In this technique, the weights of a network learned in a source task are employed as a basis to train a model in the destination task. In this way, the information learned in the source task is used in the destination task. This approach is especially beneficial when the source and target tasks are close to each other.

In our work, we study the effects of fine-tuning table detection algorithms from a distant domain (natural images from the Pascal VOC dataset~\cite{pascalvoc}) and a close domain (images of academic documents from the TableBank dataset). To this aim, we consider the following object detection algorithms.

\subsection{Object detection algorithms}
Object detection algorithms based on deep learning can be divided into two categories~\cite{Girs, yolov3}: the \emph{two-phase algorithms}, whose first step is the generation of proposals of ``interesting'' regions that are  classified using CNNs in a second step. And the \emph{one-phase algorithms} that perform detection without explicitly generating region proposals. For this work, we have employed algorithms of both types. In particular, we have used the two-phase algorithm Mask R-CNN, and the one-phase algorithms RetinaNet, SSD and YOLO. 

\paragraph{Mask R-CNN~\cite{fasterrcnn}} is currently one of the most accurate algorithm based on a two-phase approach, and the latest version of the R-CNN family. We have used a library implemented in Keras~\cite{matterport} for training models with this algorithm.

\paragraph{RetinaNet~\cite{retinaNet}} is a one-phase algorithm characterised by using the focal loss for training on a scarce set of difficult examples, and that prevents the large number of easy negatives from overwhelming the detector during training.We have used another library implemented in Keras~\cite{kerasRetina} for traning models with this algorithm.

\paragraph{SSD~\cite{ssd}} is a simple detection algorithm that completely eliminates proposal generation and encapsulates all computation in a single network. In this case, we have used the MXNET library~\cite{MxNET} for training the models.

\paragraph{YOLO~\cite{yolov3}} frames object detection as a regression problem where a single neural network predicts bounding boxes and class probabilities directly from full images in one evaluation. Although there are several versions of YOLO, the main ideas are the same for all of them. We have used the Darknet library~\cite{yolodarknet} for training models with this algorithm.

The aforementioned algorithms have been trained for detecting tables in a wide variety of document images by using the datasets presented in the following section.

\subsection{Benchmarking datasets}

For this project, we have used several datasets, see Table~\ref{datasets}. Namely, we have employed three kinds of datasets: the base datasets (which are used to train the base models), the fine-tune datasets for table detection, and the fine-tune dataset for detecting other objects in document images. The reason to consider several table detection datasets is that there are several table layouts that are highly dependent on the document type, and we want to prove that our approach can be generalised to heterogeneous document images.

\begin{table}
\centering
\begin{tabular}{lllp{4cm}}
\toprule
Datasets  & \#Train Images & \#Test Images & Type of images\\
\midrule
Pascal VOC&16,551&4,952& Natural images\\
TableBank&199,183&1,000& Academic documents\\
\midrule
ICDAR13 & 178 & 60  & Documents obtained from Google search\\
ICDAR17 &   1,200 & 400& Scientific papers   \\
ICDAR19 &   599 & 198 & Modern images \\
Invoices &  515 & 172 & Invoices\\
MarmotEn &  744 & 249  & Scientific papers \\
MarmotChi &   754 & 252 & E-books\\
UNLV &  302 & 101  & Technical reports, business letters, newspapers and magazines\\
\midrule
ICDAR17FIG &   1,200 & 400& Scientific papers  \\
ICDAR17FOR &   1,200 & 400 & Scientific papers \\
\bottomrule
\end{tabular}
\caption{Sizes of the train and test sets of the datasets}
\label{datasets}
\end{table}

\subsubsection{Base Datasets}
In this work, we have employed two datasets of considerable size for creating the base models that are later employed for fine-tuning.

\paragraph{The Pascal VOC dataset~\cite{pascalvoc}} is a popular project designed to create and evaluate algorithms for image classification, object detection and segmentation. This dataset consists of natural images which have been used for training different models in the literature. Thanks to the trend of releasing models to the public, we have employed models already trained with this dataset to apply fine-tuning from natural images to the context of table detection.
 
\paragraph{TableBank~\cite{tablebank}} is a table detection dataset built with Word and {\LaTeX} documents that contains 417K labeled images. For this project, we only employ the {\LaTeX} images (199,183 images) since the Word images contain some errors in the annotations. On the contrary to the Pascal VOC dataset, where there were available models trained for such a dataset, we have trained models for the TableBank dataset from scratch.

\subsubsection{Fine-tuning Datasets}

We have used several open datasets for fine-tuning; however, most table detection datasets only release the training set. Hence, in this project, we have divided the training sets into two sets (75\% for training  and 25\% for testing) for evaluating our approach. The dataset split are available in the project webpage, and the employed datasets are listed as follows.

\paragraph{ICDAR13~\cite{icdar13}} is one of the most famous datasets for table detection and structure recognition. This dataset is formed by documents extracted from Web pages and email messages. This dataset was prepared for a competition focused on the task of detecting tables, figures and mathematical equations from images. The dataset is comprised of PDF files which we converted to images to be used within our framework. The dataset contains 238 images in total, 178 were used for training and 60 for testing. 

\paragraph{ICDAR17~\cite{icdar17}} is a data set prepared for a competition as ICDAR13. The dataset consists of 1.600 images in total, where we can find tables, formulas and figures. The training set consists of 1,200 images, while the rest of the 400 images are used for testing. This dataset has been employed three times in our work: for the detection of tables (from now on, we will call this dataset ICDAR17), for the detection of figures (from now on, we will call this dataset ICDAR17FIG) and for the detection of formulas (from now on, we will call this dataset ICDAR17FOR).

\paragraph{ICDAR19~\cite{icdar19}} is, as in the previous cases, a dataset proposed for a competition. The dataset contains two types of images: modern documents and archival ones with various formats. In this work we have only taken the modern images (797 images in total, 599 for training and 198 for testing).

\paragraph{Invoices} is a proprietary dataset of PDF files containing invoices from several sources. The PDF files had to be converted into images. This set has 515 images in the training set and 172 in the testing set.

\paragraph{Marmot~\cite{marmot}} is a dataset that shows a great variety in language type, page layout, and table styles. Over 1,500 conference and journal papers were crawled for this dataset, covering various fields, spanning from the year 1970. to latest 2011 publications. In total, 2,000 pages in PDF format were collected. The dataset is composed of Chinese (from now on MarmotChi) and English pages (from now on MarmotEn): the MarmotChi dataset was built from over 120 e-Books with diverse subject areas provided by Founder Apabi library, and no more than 15 pages were selected from each book, this dataset contains 993 images in total, 744 were used for training and 249 for testing. And the MarmotEn dataset was crawled from the Citeseer website, this dataset contains 1,006 images in total, 754 were used for training and 252 for testing. 

\paragraph{UNLV~\cite{unlv}} is comprised of a variety of documents which includes technical reports, business letters, newspapers and magazines. The dataset contains a total of 2,889 scanned documents where only 403 documents contain a tabular region. We only used the images containing a tabular region in our experiments: 302 for training and 101 for testing.
 
Using the aforementioned algorithms and datasets, we have trained several models that have been evaluated using the following metrics.

\subsection{Performance measure}
In order to evaluate the constructed models for the different datasets, we employed the same metric used in the ICDAR19 competition for table detection~\cite{icdar19}. Considering that the ground truth bounding box is represent by GTP, and that the bounding box detected by an algorithm is represented by DTP; then, the formula for finding the overlapped region between them is given by:
$$
    IoU(GTP, DTP)  = \frac{area(GTP \bigcap DTP)}{area(GTP\bigcup DTP)}
$$
IoU(GTP, DTP) represents the overlapped region between ground truth and detected bounding boxes and its value lies between zero and one. 

Now, given a threshold $T\in [0,1]$, we define the notions of True Positive at T, TP@T, False Positive at T, FP@T, and False Negative at T, FN@T. The TP@T is the number of ground truth tables that have a major overlap $(IoU\geq T)$ with one of the detected tables. The FP@T indicates the number of detected tables that do not overlap $(IoU < T)$ with any of the ground tables. And, FN@T indicates the number of ground truth tables that do not overlap $(IoU < T)$ with any of the detected tables. From these notions, we can define the Precision at T, P@T, Recall at T, R@T,  and F1-score at T, F1@T, as follows:
$$
   P@T = \frac{TP@T}{FP@T + TP@T}
$$
$$
    R@T = \frac{TP@T}{FN@T + TP@T} 
$$
$$
    F1@T = \frac{2*TP@T}{FP@T +FN@T+ 2*TP@T} 
$$
Finally, the final score is decided by the weighted average WAvgF1 value:
$$
    WAvgF1 = \frac{0.6\times F1@0.6+0.7\times F1@0.7+0.8\times F1@0.8+0.9\times F1@0.9}{0.6+0.7+0.8+0.9}
$$
In the above formula, and since results with higher IoUs are more important than those with lower IoUs, we use IoU threshold as the weight of each F1 value to get a definitive performance score for convenient comparison. Using these metrics we have obtained the results presented in the following section.

\section{Results}\label{sec:results}
In this section, we conduct a thorough study of our approach, see Tables 2--5. Each table corresponds with the results obtained from each object detection algorithm: Table~\ref{fast} contains the results that have been obtained using Mask R-CNN; Table~\ref{retina}, the results of RetinaNet; Table~\ref{ssd}, the results of SSD; and, finally, Table~\ref{yolo} contains the results of YOLO. The tables are divided into three parts: the first row contains the results obtained for the TableBank dataset, the next 9 rows correspond with the result obtained with the models trained by fine-tuning from natural image models, and the last 9 rows correspond with the models fine-tuned from the TableBank models. All the models built in this work were trained using the default parameters in each deep learning framework, and using K80 NVIDIA GPUs.

\begin{table}
\centering
\resizebox{\textwidth}{!}{%
\begin{tabular}{llllllllllllllll}
\toprule
 & \multicolumn{3}{c}{@0.6}&\multicolumn{3}{c}{@0.7}&\multicolumn{3}{c}{@0.8}&\multicolumn{3}{c}{@0.9}&\\
&P@0.6&R@0.6&F1@0.6&P@0.7&R@0.7&F1@0.7&P@0.8&R@0.8&F1@0.8&P@0.9&R@0.9&F1@0.9&  WAvgF1 & Improvement\\
\midrule
TableBank&0.94&0.98&0.96&0.94&0.97&0.95&0.93&0.96&0.94&0.84&0.87&0.86&0.92\\
\midrule
ICDAR13&0.14&0.77&0.23&0.08&0.45&0.14&0.03&0.16&0.05&0&0&0&0.09\\
ICDAR17&0.32&0.85&0.46&0.28&0.75&0.41&0.17&0.47&0.25&0.04&0.1&0.06&0.27\\
ICDAR17FIG&0.29&0.61&0.39&0.22&0.46&0.3&0.13&0.27&0.17&0.01&0.03&0.02&0.19\\
ICDAR17FOR&0.18&0.45&0.26&0.09&0.24&0.13&0.03&0.07&0.04&0&0.01&0&0.09 \\
ICDAR19&0.6&0.64&0.62&0.48&0.51&0.5&0.24&0.25&0.25&0.02&0.02&0.02&0.31 \\
Invoices&0.38&0.56&0.45&0.28&0.42&0.34&0.16&0.23&0.19&0.02&0.03&0.02&0.22\\
MarmotEn&0.37&0.75&0.49&0.28&0.58&0.38&0.08&0.17&0.11&0&0.01&0&0.21\\
MarmotChi&0.52&0.83&0.64&0.48&0.76&0.59&0.32&0.51&0.39&0.07&0.11&0.08&0.39\\
UNLV&0.29&0.58&0.39&0.17&0.34&0.23&0.06&0.11&0.08&0.01&0.02&0.01&0.15\\

\midrule

ICDAR13&0.7&0.97&0.81&0.7&0.97&0.81&0.7&0.97&0.81&0.47&0.65&0.54&0.72 & 0.63\\
ICDAR17&0.72&0.95&0.82&0.7&0.93&0.8&0.68&0.9&0.78&0.49&0.64&0.56&0.72&0.45\\
ICDAR17FIG&0.36&0.69&0.47&0.33&0.63&0.43&0.23&0.43&0.3&0.05&0.09&0.07&0.29& 0.09\\
ICDAR17FOR&0.1&0.49&0.17&0.06&0.28&0.1&0.02&0.12&0.04&0&0.01&0&0.06 & -0.02\\
ICDAR19&0.76&0.85&0.81&0.74&0.83&0.79&0.67&0.75&0.71&0.38&0.42&0.4&0.65 & 0.34\\
Invoices&0.54&0.71&0.61&0.5&0.66&0.57&0.39&0.52&0.45&0.19&0.26&0.22&0.44 & 0.21\\
MarmotEn&0.72&0.93&0.81&0.7&0.9&0.79&0.67&0.87&0.76&0.46&0.6&0.52&0.70 & 0.48\\
MarmotChi&0.82&0.98&0.89&0.82&0.98&0.89&0.81&0.96&0.88&0.62&0.73&0.67&0.82 & 0.42\\
UNLV&0.66&0.83&0.74&0.63&0.8&0.7&0.55&0.69&0.61&0.24&0.3&0.27&0.55 & 0.39\\
\bottomrule
\end{tabular}}
\caption{Results using the Mask R-CNN algorithm}
\label{fast}
\end{table}

We start by analysing the results for the TableBank dataset, see the first row of the tables. Each model has its strenghts and weaknesses, and depending on the context we can prefer different models. The overall best model, that is the model with higher WAvgF1-score, is the Mask R-CNN model; the other three models are similar among them. If we focus on detecting as most tables as possible (R@0.6) and not detecting other artifacts as tables (P@0.6), the best model is YOLO. Finally, if we are interested in accurately detecting the regions of the tables (F1@0.9), the best model is RetinaNet. The strength of the SSD model is that it is faster than the others.

\begin{table}
\centering
\resizebox{\textwidth}{!}{%
\begin{tabular}{llllllllllllllll}
\toprule
 & \multicolumn{3}{c}{@0.6}&\multicolumn{3}{c}{@0.7}&\multicolumn{3}{c}{@0.8}&\multicolumn{3}{c}{@0.9}&\\
&P@0.6&R@0.6&F1@0.6&P@0.7&R@0.7&F1@0.7&P@0.8&R@0.8&F1@0.8&P@0.9&R@0.9&F1@0.9&  WAvgF1 & Improvement\\
\midrule
TableBank&0.98&0.86&0.92&0.98&0.86&0.92&0.97&0.85&0.91&0.94&0.82&0.87&0.90\\
\midrule
ICDAR13 & 0.56&0.58&0.57&0.56&0.58&0.57&0.56&0.58&0.57&0.34&0.35&0.35&0.50\\
ICDAR17&0.65&0.86&0.74&0.64&0.85&0.73&0.58&0.77&0.67&0.48&0.63&0.55&0.66\\
ICDAR17FIG&0.57&0.61&0.59&0.7&0.76&0.73&0.73&0.79&0.76&0.74&0.8&0.77&0.72\\
ICDAR17FOR&0.63&0.06&0.12&0.63&0.06&0.12&0.6&0.06&0.11&0.4&0.04&0.07&0.10\\
ICDAR19&0.86&0.66&0.74&0.82&0.63&0.72&0.76&0.58&0.66&0.58&0.45&0.51&0.64\\
Invoices&0.90&0.59&0.71&0.88&0.58&0.70&0.86&0.56&0.68&0.70&0.46&0.56&0.65\\
MarmotEn&0.75&0.86&0.8&0.74&0.86&0.8&0.7&0.81&0.75&0.47&0.54&0.5&0.69\\
MarmotChi&0.78&0.85&0.81&0.75&0.81&0.78&0.73&0.79&0.75&0.5&0.54&0.52&0.70\\
UNLV &0.81&0.83&0.82&0.79&0.81&0.80&0.76&0.77&0.76&0.61&0.63&0.62&0.73\\
\midrule
ICDAR13&0.83&0.77&0.8&0.79&0.74&0.77&0.76&0.71&0.73&0.72&0.68&0.7&0.74&0.24\\
ICDAR17&0.92&0.87&0.89&0.92&0.87&0.89&0.89&0.84&0.86&0.79&0.75&0.77&0.84&0.18\\
ICDAR17FIG&0.76&0.79&0.77&0.74&0.77&0.76&0.72&0.75&0.74&0.64&0.66&0.65&0.72&0.001\\
ICDAR17FOR&0.26&0.35&0.3&0.24&0.32&0.27&0.19&0.26&0.22&0.08&0.1&0.09&0.20&0.11\\
ICDAR19&0.91&0.74&0.82&0.87&0.81&0.79&0.81&0.67&0.73&0.68&0.56&0.61&0.72&0.08\\
Invoices&0.92&0.59&0.72&0.92&0.59&0.71&0.87&0.55&0.68&0.74&0.47&0.58&0.66&0.01\\
MarmotEn&0.93&0.86&0.9&0.92&0.86&0.89&0.91&0.84&0.87&0.78&0.73&0.75&0.84&0.14\\
MarmotChi&0.87&0.87&0.87&0.85&0.85&0.85&0.83&0.83&0.83&0.69&0.7&0.69&0.80&0.10\\
UNLV&0.81&0.83&0.82&0.79&0.8&0.8&0.75&0.77&0.76&0.63&0.64&0.64&0.74&0.01\\
\bottomrule
\end{tabular}}
\caption{Results using the RetinaNet algorithm}
\label{retina}
\end{table}

\begin{figure}
    \centering
    \includegraphics[scale=0.1]{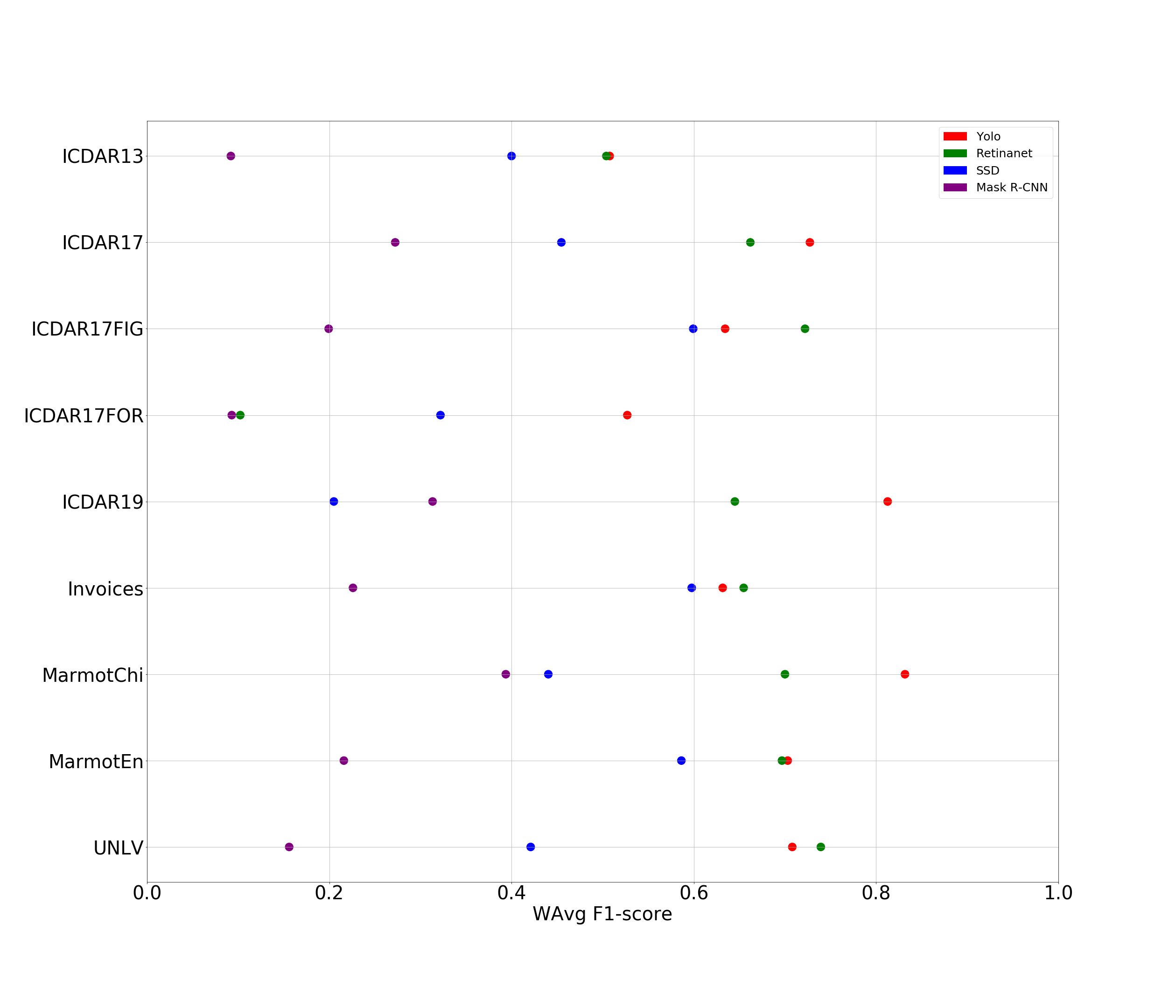}
    \caption{Dispersion diagram using fine-tuning from natural images}
    \label{fig:das}
\end{figure}

\begin{figure}
    \centering
    \includegraphics[scale=0.1]{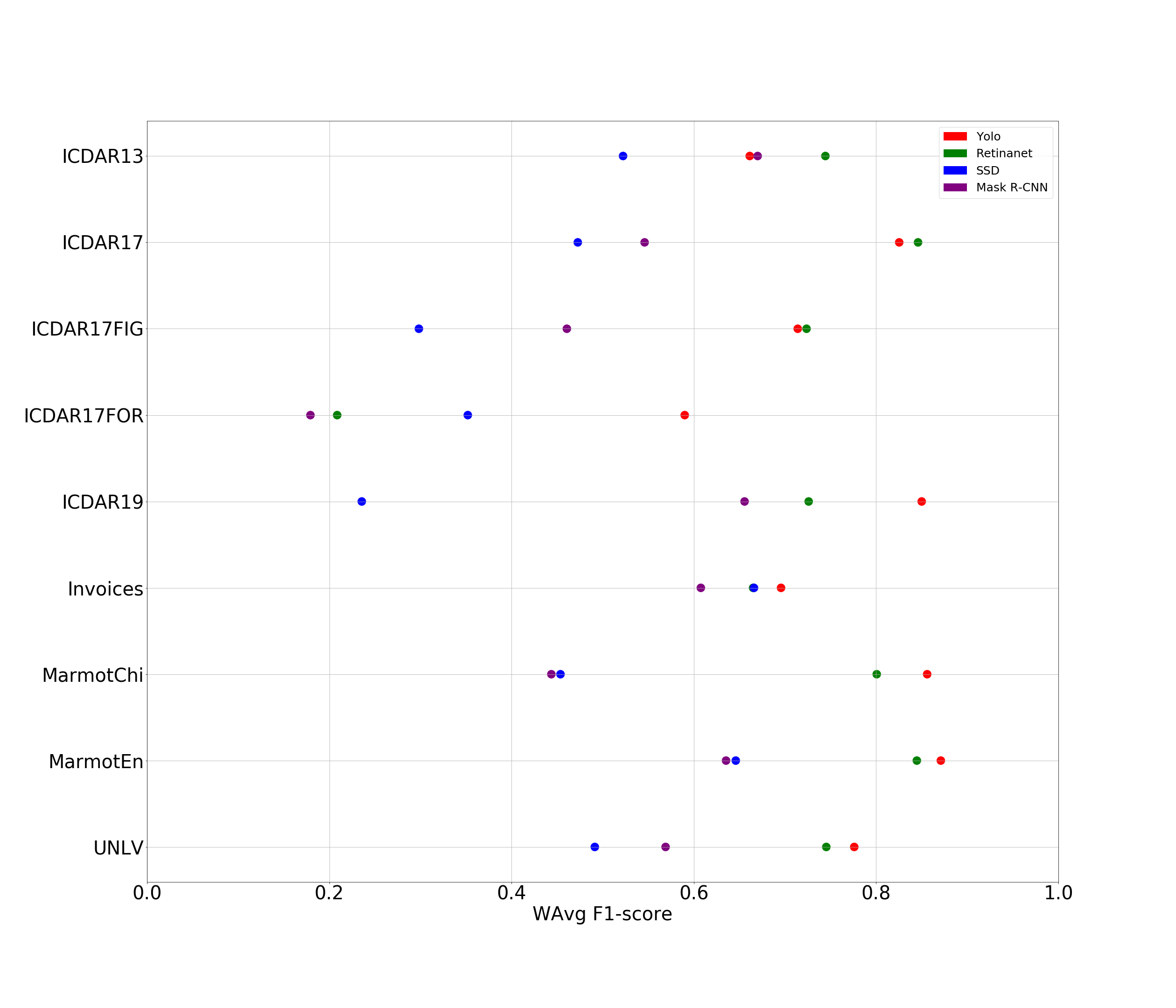}
    \caption{Dispersion diagram using fine-tuning dataset from the TableBank}
    \label{fig:dasTrans}
\end{figure}

Let us focus now on the table detection datasets. In the case of models fine-tuned using natural images, the algorithms that stand out are RetinaNet and YOLO, see Tables~\ref{fast} to 5 and Figure~\ref{fig:das}. Similary, the models that achieve higher accuracies when fine-tuning from the TableBank dataset are YOLO and RetinaNet, see Figure~\ref{fig:dasTrans}. As can be seen in Tables 2 to 5, fine-tuning from a close domain produce more accurate models that fine-tuning from an unrelated domain. However, the effects on each algorithm and dataset greatly differ. The algorithm that is more considerably boosted for table detection is Mask R-CNN, that improves up to a 60\% in some cases and 42\% in mean. In the case of RetinaNet, in mean it improves by 11\%, YOLO a 9\%, and SSD is the one with the least improvement, only a 5\%.

Finally, if we consider the results for the datasets containing figures and formulas, the improvement is not as remarkable as in the detection of tables. In this case, the algorithm that takes a bigger advantage of this technique is YOLO, since in both cases it improves up to a 10\%. In the case of RetinaNet, it improves the detection of formulas by 10\%, while that of figures barely improves. And in the case of SSD and Mask R-CNN, they are the ones with the least improvement and even have some penalty.

\begin{table}
\centering
\resizebox{\textwidth}{!}{%
\begin{tabular}{llllllllllllllll}
\toprule
 & \multicolumn{3}{c}{@0.6}&\multicolumn{3}{c}{@0.7}&\multicolumn{3}{c}{@0.8}&\multicolumn{3}{c}{@0.9}&\\
&P@0.6&R@0.6&F1@0.6&P@0.7&R@0.7&F1@0.7&P@0.8&R@0.8&F1@0.8&P@0.9&R@0.9&F1@0.9& WAvgF1 & Improvement\\
\midrule
TableBank&0.96&0.97&0.96&0.94&0.95&0.95&0.92&0.92&0.92&0.82&0.82&0.82&0.90\\
\midrule
ICDAR13 &0.54&0.68&0.6&0.44&0.55&0.49&0.38&0.48&0.43&0.15&0.19&0.17&0.40\\
ICDAR17&0.49&0.71&0.58&0.41&0.59&0.48&0.34&0.49&0.4&0.22&0.32&0.4&0.45\\
ICDAR17FIG&0.7&0.8&0.75&0.68&0.77&0.72&0.61&0.69&0.65&0.34&0.38&0.36&0.59\\
ICDAR17FOR&0.44&0.64&0.52&0.34&0.49&0.4&0.19&0.27&0.22&0.03&0.05&0.22&0.32\\
ICDAR19&0.31&0.35&0.33&0.23&0.26&0.25&0.18&0.2&0.19&0.1&0.11&0.1&0.20\\
Invoices&0.87&0.84&0.85&0.8&0.78&0.79&0.63&0.61&0.62&0.27&0.26&0.26&0.59\\
MarmotEn &0.67&0.76&0.71&0.63&0.71&0.67&0.6&0.67&0.63&0.38&0.43&0.4&0.58\\
MarmotChi &0.57&0.7&0.63&0.48&0.60&0.53&0.36&0.45&0.4&0.25&0.31&0.28&0.44\\
UNLV &0.66&0.64&0.65&0.6&0.58&0.59&0.45&0.43&0.44&0.12&0.11&0.12&0.42\\

\midrule

ICDAR13 &0.62&0.68&0.65&0.62&0.68&0.65&0.5&0.55&0.52&0.32&0.35&0.34&0.52&0.12\\
ICDAR17&0.55&0.71&0.62&0.46&0.60&0.52&0.42&0.54&0.47&0.30&0.39&0.34&0.47&0.01\\
ICDAR17FIG&0.27&0.77&0.41&0.26&0.72&0.38&0.21&0.58&0.31&0.10&0.28&0.15&0.29&-0.3\\
ICDAR17FOR&0.5&0.74&0.6&0.41&0.6&0.49&0.29&0.42&0.34&0.07&0.11&0.09&0.35&0.03\\
ICDAR19&0.35&0.35&0.35&0.28&0.27&0.28&0.23&0.22&0.23&0.13&0.12&0.13&0.23&0.03\\
Invoices&0.91&0.86&0.89&0.87&0.81&0.84&0.71&0.67&0.69&0.37&0.35&0.36&0.66&0.07\\
MarmotEn&0.71&0.75&0.73&0.69&0.73&0.71&0.66&0.70&0.68&0.49&0.52&0.51&0.64&0.06\\
MarmotChi&0.61&0.67&0.64&0.50&0.55&0.52&0.42&0.46&0.44&0.28&0.31&0.29&0.45&0.01\\
UNLV &0.72&0.66&0.69&0.66&0.61&0.63&0.5&0.45&0.47&0.28&0.26&0.27&0.49&0.07\\
\bottomrule
\end{tabular}}
\caption{Results using the SSD algorithm}
\label{ssd}
\end{table}

\begin{table}
\centering
\resizebox{\textwidth}{!}{%
\begin{tabular}{lllllllllllllll}
\toprule
 & \multicolumn{3}{c}{@0.6}&\multicolumn{3}{c}{@0.7}&\multicolumn{3}{c}{@0.8}&\multicolumn{3}{c}{@0.9}&\\
&P@0.6&R@0.6&F1@0.6&P@0.7&R@0.7&F1@0.7&P@0.8&R@0.8&F1@0.8&P@0.9&R@0.9&F1@0.9& WavgF1 & Improvement\\
\midrule
TableBank&0.98&0.99&0.98&0.98&0.99&0.98&0.96&0.97&0.96&0.74&0.75&0.75&0.90\\
\midrule
ICDAR13 &0.92&0.58&0.6&0.61&0.61&0.61&0.57&0.55&0.56&0.31&0.32&0.32&0.50\\
ICDAR17&0.9&0.94&0.92&0.88&0.93&0.9&0.78&0.82&0.8&0.39&0.41&0.4&0.72\\
ICDAR17FIG &0.88&0.84&0.86&0.85&0.82&0.84&0.75&0.72&0.74&0.23&0.22&0.23&0.63\\
ICDAR17FOR &0.9&0.85&0.87&0.82&0.77&0.79&0.54&0.5&0.52&0.1&0.09&0.1&0.52\\
ICDAR19&0.95&0.91&0.93&0.94&0.9&0.92&0.89&0.85&0.87&0.61&0.58&0.6&0.81\\
Invoices &0.89&0.87&0.88&0.84&0.82&0.83&0.7&0.68&0.69&0.26&0.26&0.26&0.63\\
MarmotEn &0.9&0.96&0.93&0.87&0.93&0.9&0.76&0.81&0.78&0.32&0.34&0.33&0.70\\
MarmotChi &0.95&0.96&0.96&0.93&0.94&0.94&0.88&0.89&0.89&0.61&0.62&0.61&0.83\\
UNLV &0.91&0.95&0.93&0.88&0.91&0.89&0.73&0.76&0.74&0.39&0.4&0.39&0.70\\
\midrule

ICDAR13&1&0.65&0.78&0.95&0.61&0.75&0.9&0.58&0.71&0.6&0.39&0.47&0.66&0.15\\
ICDAR17&0.94&0.94&0.94&0.93&0.94&0.93&0.89&0.89&0.89&0.61&0.62&0.61&0.82&0.09\\
ICDAR17FIG &0.91&0.83&0.87&0.89&0.82&0.86&0.83&0.76&0.8&0.44&0.4&0.42&0.71&0.07\\
ICDAR17FOR&0.94&0.85&0.89&0.88&0.79&0.83&0.65&0.59&0.62&0.19&0.17&0.18&0.59&0.06\\
ICDAR19&0.95&0.95&0.95&0.94&0.94&0.94&0.9&0.9&0.9&0.68&0.68&0.68&0.85&0.04\\
Invoices &0.9&0.89&0.89&0.87&0.85&0.86&0.76&0.74&0.75&0.39&0.39&0.39&0.69&0.06\\
MarmotEn &0.95&0.97&0.96&0.95&0.97&0.96&0.92&0.93&0.93&0.68&0.69&0.69&0.87&0.16\\
MarmotChi &0.97&0.93&0.95&0.96&0.93&0.94&0.92&0.89&0.91&0.69&0.67&0.68&0.85&0.02\\
UNLV&0.93&0.95&0.94&0.92&0.94&0.93&0.83&0.85&0.84&0.48&0.49&0.49&0.77&0.06\\
\bottomrule
\end{tabular}}
\caption{Results using the YOLO algorithm}
\label{yolo}
\end{table}

As we have shown in this section, fine-tuning from the TableBank dataset can boosten table detection models. However, there is not a model that outperforms the rest, see Figures \ref{fig:das} and \ref{fig:dasTrans}. Therefore, we have released a set of tools to employ the trained models, and also employ them for constructing models using fine-tuning on custom datasets.

\section{Tools for table detection}\label{sec:tools}
Using one of the generated detection model with new images is usually as simple as invoking a command with the path of the image (and, probably, some additional parameters). However, this requires the installation of several libraries and the usage of a command line interface; and, this might be challenging for non-expert users. Therefore, it is important to create simple and intuitive interfaces that might be employed by different kinds of users; otherwise, they will not be able to take advantage of the object detection models.

To disseminate our detection models, we have created a set of Jupyter notebooks, that allows users to detect tables in their images. Jupyter notebooks~\cite{jupyter} are documents for publishing code, results and explanations in a form that is both readable and executable; and, they have been widely adopted across multiple disciplines, both for their usefulness in keeping a record of data analyses, and also for allowing reproducibility. The drawback of Jupyter notebooks is that they require the installation of several libraries. Such a problem has been overcome in our case by providing our notebooks in Google Colaboratory~\cite{colab}, a free Jupyter notebook environment that requires no setup and runs entirely in the cloud avoiding the installation of libraries in the local computer. The notebooks are available at \href{https://github.com/holms-ur/fine-tuning}{https://github.com/holms-ur/fine-tuning}.

In addition, in the same project webpage, due to the heterogeneity of document images containing tables, we have provided all the weights, configuration files, and necessary instructions to fine-tune any of the detection models created in this work to custom datasets containing tables.

\section{Conclusion and Further work}
In this work, we have shown the benefits of using fine-tuning from a close domain in the context of table detection. In addition to the accuracy improvement, this approach avoids overfitting and solves the problem of having a small dataset. Moreover, we can highlight that apart from the Mask R-CNN algorithm, other algorithms such as YOLO and RetinaNet can achieve a good performance in the table detection task.

Since table detection is the first step towards table analysis, we plan to use this work as a basis for determining the internal structure of tables, and, eventually, extracting the semantics from table contents. Moreover, we are also interested in extending these methods to detect forms in document images.

\bibliographystyle{splncs04}
\bibliography{biblio}

\begin{thebibliography}{10}
\providecommand{\url}[1]{\texttt{#1}}
\providecommand{\urlprefix}{URL }
\providecommand{\doi}[1]{https://doi.org/#1}

\bibitem{matterport}
Abdulla, W.: {Mask R-CNN for object detection and instance segmentation on
  Keras and TensorFlow} (2017), \url{https://github.com/matterport/Mask_RCNN}

\bibitem{yolodarknet}
Alexey, A.B.: {YOLO darknet} (2018), \url{https://github.com/AlexeyAB/darknet}

\bibitem{Cesari02}
Cesari, F., et~al.: {Trainable Table Location in Document Images}. In: 16th
  International Conference on Pattern Recognition. ICPR'02, vol.~3, p. 30236.
  ACM (2002)

\bibitem{MxNET}
Chen, T., et~al.: {MXNet: A Flexible and Efficient Machine Learning Library for
  Heterogeneous Distributed Systems}. CoRR  \textbf{abs/1512.01274} (2015),
  \url{http://arxiv.org/abs/1512.01274}

\bibitem{colab}
{Colaboratory team}: Google colaboratory (2017),
  \url{https://colab.research.google.com}

\bibitem{Costa09}
{Costa e Silva}, A.: {Learning Rich Hidden Markov Models in Document Analysis:
  Table Location}. In: 10th International Conference on Document Analysis and
  Recognition. pp. 843--847. ICDAR'10, IEEE (2009)

\bibitem{Couasnon14}
Coüasnon, B., Lemaitre, A.: Handbook of Document Image Processing and
  Recognition, chap. Recognition of Tables and Forms, pp. 647--677. Springer
  International Publishing (2014)

\bibitem{Embley06}
Embley, D.W., et~al.: Table-processing paradigms: a research survey.
  International Journal Document Analysis and Recognition  \textbf{8}(2--3),
  647--677 (2006)

\bibitem{pascalvoc}
Everingham, M., et~al.: {The Pascal Visual Object Classes Challenge: A
  Retrospective}. International Journal of Computer Vision  \textbf{111}(1),
  98--136 (2015)

\bibitem{icdar17}
Gao, L., Yi, X., Jiang, Z., Hao, L., Tang, Z.: {ICDAR2017 competition on page
  object detection}. In: 14th IAPR International Conference on Document
  Analysis and Recognition. pp. 1417--1422. ICDAR'17 (2017)

\bibitem{Gilani17}
Gilani, A., et~al.: {Table Detection using Deep Learning}. In: 14th
  International Conference on Document Analysis and Recognition. pp. 771--776.
  ICDAR'17, IEEE (2017)

\bibitem{Girs}
Girshick, R., et~al.: {Accurate Object Detection and Semantic Segmentation}.
  In: 2014 IEEE Computer Society Conference on Computer Vision and Pattern
  Recognition. pp. 580--587. CVPR'14, IEEE (2014)

\bibitem{icdar13}
Gobel, M.C., Hassan, T., Oro, E., Orsi, G.: {ICDAR2013 Table Competition}. In:
  12th ICDAR Robust Reading Competition. pp. 1449--1453. ICDAR'13, IEEE (2013)

\bibitem{Goodfellow16}
Goodfellow, I., Bengio, Y., Courville, A.: Deep Learning. MIT Press (2016),
  \url{http://www.deeplearningbook.org}

\bibitem{Hao16}
Hao, L., et~al.: {A table detection method for pdf documents based on
  convolutional neural networks}. In: 12th International Workshop on Document
  Analysis Systems. pp. 287--292. DAS'16, IEEE (2016)

\bibitem{Hirayama95}
Hirayama, Y.: {A method for table structure analysis using DP matching}. In:
  3rd International Conference on Document Analysis and Recognition. pp.
  583--586. ICDAR'95, IEEE (1995)

\bibitem{Huang19}
Huang, Y., et~al.: {A YOLO-based Table Detection Method}. In: 15th
  International Conference on Document Analysis and Recognition. ICDAR'19
  (2019)

\bibitem{marmot}
{Institute of Computer Science and Techonology of Peking University and
  Institute of Digital Publishing of Founder R\&D Center, China}: Marmot
  dataset for table recognition (2011),
  \url{http://www.icst.pku.edu.cn/cpdp/sjzy/index.htm}

\bibitem{Jianying99}
Jianying, H., et~al.: {Medium-independent table detection}. In: Document
  Recognitionand Retrieval VII. vol.~3967, pp. 583--586. International Society
  for Optics and Photonics (1999)

\bibitem{Kasar13}
Kasar, T., et~al.: {Learning to Detect Tables in Scanned Document Images Using
  Line Information}. In: 12th International Conference on Document Analysis and
  Recognition. pp. 1185--1189. ICDAR'13, IEEE (2013)

\bibitem{Kerwat18}
Kerwat, M., George, R., Shujaee, K.: {Detecting Knowledge Artifacts in
  Scientific Document Images - Comparing Deep Learning Architectures}. In: 5th
  International Conference on Social Networks Analysis, Management and
  Security. pp. 147--152. SNAMS'18, IEEE (2018)

\bibitem{jupyter}
Kluyver, T., et~al.: Jupyter notebooks --- a publishing format for reproducible
  computational workflows. In: 20th International Conference on Electronic
  Publishing. pp. 87--90. IOS Press (2016)

\bibitem{tablebank}
Li, M., et~al.: {TableBank: Table Benchmark for Image-based Table Detection and
  Recognition}. CoRR  \textbf{abs/1903.01949} (2019),
  \url{http://arxiv.org/abs/1903.01949}

\bibitem{kerasRetina}
Lin, T., Goyal, P., Girshick, R., He, K., Dollár., P.: Keras retinanet (2017),
  \url{https://github.com/fizyr/keras-retinanet}

\bibitem{retinaNet}
Lin, T.Y., et~al.: {Focal Loss for Dense Object Detection}. In: 16th
  International Conference on Computer Vision. pp. 2999--3007. ICCV'17 (2017)

\bibitem{ssd}
Liu, W., et~al.: {SSD: Single Shot MultiBox Detectors}. In: 14th European
  Conference on Computer Vision. ECCV'16, vol.~9905, pp. 21--37 (2016)

\bibitem{Borges17}
Oliveira, D.A.B., Viana, M.P.: {Fast CNN-based document layout analysis}. In:
  14th International Conference on Computer Vision Workshops. pp. 1173--1180.
  ICCVW'17, IEEE (2017)

\bibitem{Oro09}
Oro, E., Ruffolo, M.: {PDF-TREX: An approach for recognizing and extracting
  tables from PDF documents}. In: 10th International Conference on Document
  Analysis and Recognition. pp. 906--910. ICDAR'09, IEEE (2009)

\bibitem{Razavian14}
Razavian, A.S., et~al.: {CNN features off-the-shelf: An astounding baseline for
  recognition}. In: 27th Conference on Computer Vision and Pattern Recognition
  Workshops. pp. 512--519. CVPRW'14 (2014)

\bibitem{yolov3}
Redmon, J., Farhadi, A.: {YOLOv3: An Incremental Improvement}. CoRR
  \textbf{abs/1804.02767} (2018), \url{http://arxiv.org/abs/1804.02767}

\bibitem{fasterrcnn}
Ren, S., He, K., Girshick, R., Sun, J.: {Faster R-CNN: Towards Real-Time Object
  Detection with Region Proposal Networks}. Advances in Neural Information
  Processing Systems  \textbf{28},  91--99 (2015)

\bibitem{Pyimagesearch}
Rosebrock, A.: {Deep Learning for Computer Vision with Python}. PyImageSearch
  (2018), \url{https://www.pyimagesearch.com/}

\bibitem{ILSVRC15}
Russakovsky, O., et~al.: {ImageNet Large Scale Visual Recognition Challenge}.
  International Journal of Computer Vision  \textbf{115}(3),  211--252 (2015)

\bibitem{Schreiber17}
Schreiber, S., et~al.: {DeepDeSRT: Deep Learning for Detection and Structure
  Recognition of Tables in Document Images}. In: 14th International Conference
  on Document Analysis and Recognition. pp. 1162--1167. ICDAR'17, IEEE (2017)

\bibitem{unlv}
Shahab, A., Shafait, F., Kieninger, T., Dengel, A.: {An open approach towards
  the benchmarking of table structure recognition systems}. In: 9th IAPR Int.
  Workshop on Document Analysis Systems. pp. 113--120. DAS'10 (2010)

\bibitem{Siddiqui18}
Siddiqui, S.A., et~al.: {DeCNT: Deep Deformable CNN for Table Detection}. IEEE
  Access  \textbf{6},  74151--74161 (2018)

\bibitem{icdar19}
Suen, C.Y., et~al.: {ICDAR2019 Table Competition} (2019),
  \url{http://icdar2019.org/}

\bibitem{Zanibbi04}
Zanibbi, R., Blostein, D., Cordy, J.R.: {A survey of table recognition}.
  Document Analysis and Recognition  \textbf{7}(1),  1--16 (2004)

\end{thebibliography}

\end{document}